\pdfoutput=1

\documentclass[11pt]{article}

\usepackage[]{acl}

\usepackage{times}
\usepackage{latexsym}

\usepackage[T1]{fontenc}

\usepackage[utf8]{inputenc}

\usepackage{microtype}

\usepackage{inconsolata}

\usepackage{graphicx}
\usepackage[nolist]{acronym}
\usepackage{multirow}
\usepackage{url}
\usepackage{graphicx}
\usepackage{arydshln}
\usepackage{listings}

\usepackage{tikz}
\newcommand{\tikzxmark}{%
\tikz[scale=0.23] {
    \draw[line width=0.7,line cap=round] (0,0) to [bend left=6] (1,1);
    \draw[line width=0.7,line cap=round] (0.2,0.95) to [bend right=3] (0.8,0.05);
}}
\newcommand{\tikzcmark}{%
\tikz[scale=0.23] {
    \draw[line width=0.7,line cap=round] (0.25,0) to [bend left=10] (1,1);
    \draw[line width=0.8,line cap=round] (0,0.35) to [bend right=1] (0.23,0);
}}

\begin{acronym}
    \acro{rq}[RQ]{research question}
    \acroplural{rq}[RQs]{research questions}
    \acro{nlp}[NLP]{natural language processing}
    \acro{llm}[LLM]{large language model}
    \acro{lora}[LoRA]{low-rank adaptation}
    \acroplural{llm}[LLMs]{large language models}
    \acroplural{plm}[PLMs]{pre-trained language models}
    \acro{ai}[AI]{artificial intelligence}
    \acro{qa}[QA]{Question answering}
    \acro{kbqa}[KBQA]{question answering over knowledge bases}
    \acro{nlg}[NLG]{Natural language generation}
    \acro{kg}[KG]{knowledge graph}
    \acroplural{kg}[KGs]{knowledge graphs}
\end{acronym}

\title{MedREQAL: Examining Medical Knowledge Recall of Large \\ Language Models via Question Answering}

\author{Juraj Vladika, Phillip Schneider, Florian Matthes \\
  Technical University of Munich \\
  TUM School of CIT \\
  Garching, Germany \\
  \texttt{\{juraj.vladika, phillip.schneider, matthes\}@tum.de}}

\begin{document}
\maketitle
\begin{abstract}
In recent years, Large Language Models (LLMs) have demonstrated an impressive ability to encode knowledge during pre-training on large text corpora. They can leverage this knowledge for downstream tasks like question answering (QA), even in complex areas involving health topics. Considering their high potential for facilitating clinical work in the future, understanding the quality of encoded medical knowledge and its recall in LLMs is an important step forward. In this study, we examine the capability of LLMs to exhibit medical knowledge recall by constructing a novel dataset derived from systematic reviews -- studies synthesizing evidence-based answers for specific medical questions. Through experiments on the new MedREQAL dataset, comprising question-answer pairs extracted from rigorous systematic reviews, we assess six LLMs, such as GPT and Mixtral, analyzing their classification and generation performance. Our experimental insights into LLM performance on the novel biomedical QA dataset reveal the still challenging nature of this task.

\end{abstract}
\section{Introduction}
The field of \ac{nlp} has been transformed with the advent of pre-trained \acp{llm}. During their process of pre-training to predict the next token on massive amounts of text data, these models learn and construct an internalized representation of world knowledge \cite{zhang-etal-2023-large}. A popular domain of application of \acp{llm} is healthcare, where they have the potential to democratize medical knowledge and facilitate access to healthcare, but also introduce risks of misinformation and lack of transparency \cite{clusmann2023future}. Recent work has hinted at the fact that \acp{llm} encode clinical knowledge rather well \cite{singhal2023large}. 


In the medical world, \textit{systematic reviews } are overview studies that synthesize the best available studies on a clearly defined medical research question. In them, the studies on a topic are analyzed, critically appraised, and their interpretations summarized by experts into a refined evidence-based conclusion \cite{pollock2018systematic}. Considering they synthesize the best available knowledge for a medical question into a concise answer, we see systematic reviews as a very well-suited proxy for testing the level of \textit{knowledge recall} in \acp{llm}. 

When prompted in a zero-shot setting, instruction-tuned \acp{llm} also tend to generate a conclusion addressing the question based on their recalled knowledge from diverse sources. To examine how well the \acp{llm} perform medical knowledge recall, we constructed a question-answering (QA) dataset originating from systematic reviews and evaluated the performance of six \acp{llm}, probing their medical knowledge recall. Based on experiments conducted on the newly constructed MedREQAL dataset, we analyze both the classification and generation performance, discussing insights about individual capabilities and limitations.

Our contributions include: (1) a novel dataset of biomedical question-answer pairs originating from rigorous systematic reviews, (2) experiments testing the zero-shot medical knowledge recall of six \acp{llm}, and (3) deeper analysis of the dataset and experimental results. To ensure reproducibility, we provide the dataset and code in a public repository.\footnote{\url{https://github.com/jvladika/MedREQAL}}

\section{Related Work}
Medicine is a common domain of application for NLP tasks \cite{thirunavukarasu2023large}. It is marked by the highly complex biomedical language and terminology. Previous studies have highlighted the potential of LLMs in medical knowledge recall and exposed key gaps indicating the importance of further method development for creating safe and effective LLMs for health-related applications \cite{lievin2022can,singhal2023large}. 

Biomedical question answering (BQA) can be split into four main categories: scientific, clinical, consumer, and examination \cite{jin2022biomedical}. Our dataset best fits into the scientific category, where questions and answers originate directly from biomedical research publications, and the most well-known datasets are PubMedQA \cite{jin-etal-2019-pubmedqa} and BioASQ \cite{bioasqqa}. While the usual QA setting works with provided documents, recent efforts have started advancing the open-domain QA setting, where evidence first has to be discovered in order to answer the question \cite{jin2021disease, vladika-matthes-2024-comparing, vladika2024improving, tian2024opportunities}.

Also related is the task of scientific fact-checking, focusing on assessing the veracity of claims based on relevant scientific evidence \cite{wadden-etal-2020-fact, vladika-matthes-2023-scientific}. The most similar dataset to MedREQAL is HealthFC \cite{vladika-etal-2024-healthfc-verifying}, which also uses systematic reviews as its evidence source for answers but focuses on popular health inquiries and retells the reviews in lay language. Systematic reviews are most commonly associated in NLP with the task of their automatic construction \cite{marshall2019toward, van2023artificial}, which is a highly complex task eliciting advanced biomedical knowledge and refined reasoning skills. To the best of our knowledge, we introduce the first BQA dataset directly generated from systematic reviews, all adhering rigorously to a predefined research protocol.

\section{Dataset}
\paragraph{Data Source}
Our data originates from systematic reviews conducted by the Cochrane Collaboration. Systematic reviews synthesize the best available evidence on a clearly defined medical research question and then provide a concise evidence-based conclusion \cite{pollock2018systematic}. The Cochrane Collaboration is a global charitable organization formed with the aim of improving evidence-based healthcare decision-making through systematic reviews of the effects of healthcare interventions \cite{henderson2010write}. 
Cochrane Reviews are done by 30,000 volunteer medical experts, who follow a strict methodology and a highly structured format (with same seven sections in every abstract) for easier publishing in the Cochrane database \cite{cumpston2022chapter}.   
We deem that the focus on important healthcare interventions, wide domain coverage, standardized structure, and rigorous process followed by medical experts to construct these reviews are all factors that made them a highly suitable source for the construction of a novel biomedical question-answering dataset.

\begin{table}[t]
\centering

\begin{tabular}{p{\columnwidth}}
\\
\hline
\textbf{Objective:} \small To assess the effects of listening to music on sleep in adults with insomnia and to assess the influence of specific variables that may moderate the effect.
 \\ 

\textbf{Generated question:} \small Can listening to music improve sleep in adults with insomnia?\\

\textbf{Conclusion:} \small The findings of this review provide evidence that music may be effective for improving subjective sleep quality in adults with symptoms of insomnia. (...)
 \\

\textbf{Verdict:}  \textcolor{teal}{\textbf{Supported}} \small \cite{Jespersen2022-uy} \\
\hline
\textbf{Objective:} \small To assess the effects of alpha-lipoic acid as a disease-modifying agent in people with diabetic peripheral neuropathy.
 \\

\textbf{Generated question:} \small Does alpha-lipoic acid have a disease-modifying effect on diabetic peripheral neuropathy? \\

\textbf{Conclusion:} \small Our analysis suggests that ALA probably has little or no effect on neuropathy symptoms or adverse events at six months, and may have little or no effect on impairment at six months. (...)
 \\

\textbf{Verdict:}  \textcolor{red}{\textbf{Refuted}} \small \citep{Baicus2024-sf} \\
\hline
\end{tabular}

\caption{\label{tab:claims}Example of two instances from the \textsc{MedREQAL} dataset. The question is generated from the original objective. The original conclusion is used for the answer generation task and the verdict is used for the classification task.}
\end{table}

\paragraph{Dataset Construction}

To construct the dataset, we first scraped the abstracts of all the available Cochrane systematic reviews in the database PubMed, published from 2018 to 2023,\footnote{\url{https://pubmed.ncbi.nlm.nih.gov/?term=\%22Cochrane+Database+syst+rev\%22\%5BJournal\%5D&filter=years.2018-2023}} using the Python library Beautiful Soup.
The final QA dataset consists of (1) questions, (2) labels, and (3) long answers. Questions were generated by leveraging the \textit{Objectives} section present in every review. The objective text (usually one declarative sentence) was given to an LLM and instructed to form a question from it using GPT-3.5 (Turbo-0125) (see Table~\ref{tab:data_prompts} in Appendix for all prompt details). Long answers are original words taken directly from the \textit{Authors' conclusions} section present in each review. On top of these answers, we also generated a discreet label for each of these conclusions, which was one of the three: \textsc{Supported}, \textsc{Refuted}, \textsc{Not enough information}. The motivation for this is two-fold: (1) better alignment with existing comparable biomedical fact-checking and QA datasets that have these three labels, and (2) this allows for an easier evaluation of experiments with classification metrics like F1. The generation of these labels was also done with GPT-3.5 (see prompt in Table~\ref{tab:data_prompts} in Appendix). Ultimately, the generated questions were sorted into various health areas using a classification prompt featuring 13 options, with a fourteenth category created afterward to consolidate questions concerning the renal and urinary systems. The classification process was performed using the same GPT-3.5 model with the temperature parameter set to 0. Two authors evaluated the generated questions and labels by randomly selecting 100 examples ($\sim$4\% of all questions) and found 94\% of questions and 92\% of labels to be correct, which we deem satisfying. The most common error in question generation was generating a question about the study itself (e.g., "\textit{Did the study analyze interventions to (...)?}"), while the most common label error was mislabeling a refuting verdict with a low level of evidence as "not enough information".

\paragraph{Dataset Description}

The constructed dataset comprises a total of 2,786 generated questions, each averaging 16.6 words in length with a std. dev. of 5.4. The 
distribution of labels shows that a significant portion of the medical questions (2057) lack sufficient information for a conclusive verdict, while 543 questions are supported by evidence, and 186 are refuted based on available data. While the proportion of \textsc{NEI} answers seems high (74\%) as opposed to supported \& refuted (26\%), this is the result of strict guidelines followed by Cochrane reviewers for giving a conclusive verdict. A similar distribution was found by \citet{howick2022most}, where 2,428 Cochrane reviews were manually analyzed, and the authors note "that only 26\% of interventions had effects that were supported by moderate quality evidence according to GRADE\footnote{\url{https://bestpractice.bmj.com/info/toolkit/learn-ebm/what-is-grade/}}". 

\begin{figure}[htpb]
  \centering
  \includegraphics[width=0.49\textwidth]{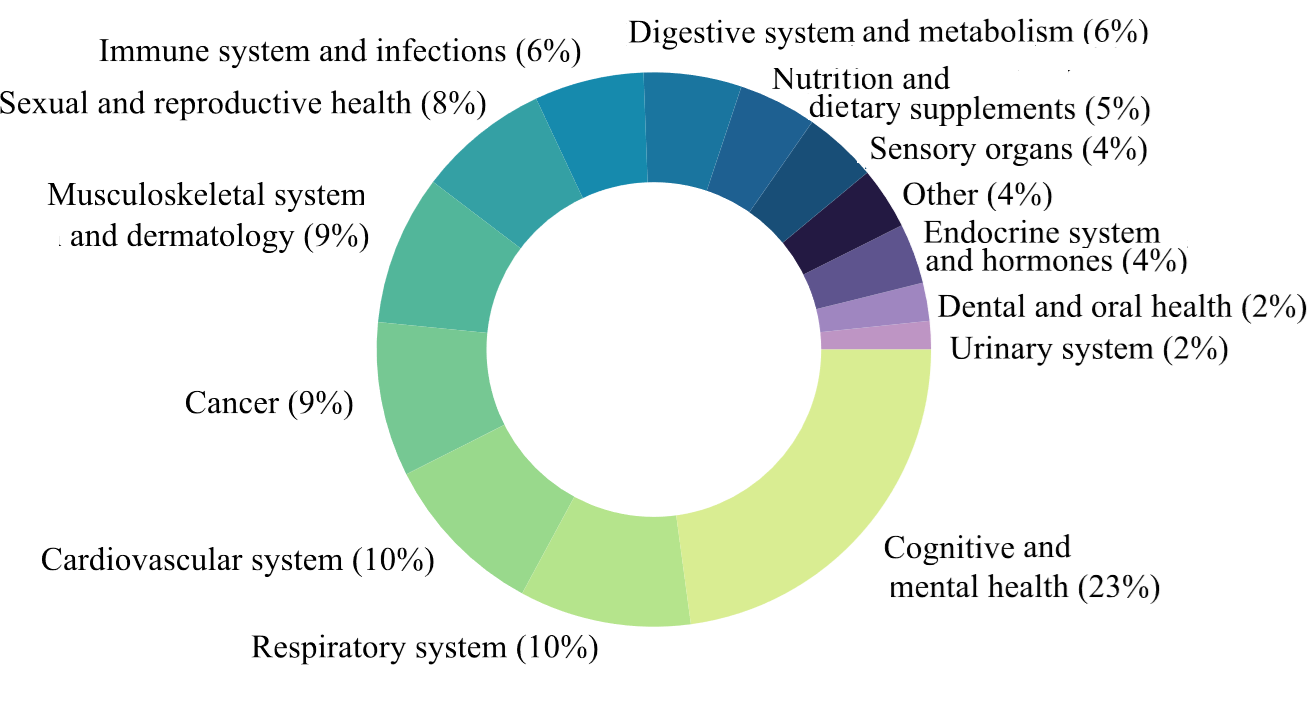}
  \caption{Percentage distribution of classified health areas in the \textsc{MedREQAL} dataset.}
\label{fig:health-categories}
\end{figure}

As visualized in Figure~\ref{fig:health-categories}, the distribution of questions across the health categories in our dataset exhibits a relatively balanced distribution, with most topics having a comparable percentage, indicating an even spread and thus contributing to the dataset's quality. Being the majority class, cognitive and mental health comprises the largest proportion at 23\%. Respiratory, cardiovascular, and cancer-related inquiries each represent approximately every tenth question, highlighting significant attention to these vital areas. Other important areas include musculoskeletal health, sexual and reproductive health, and immune system-related questions. The dataset also covers questions concerning digestive health, nutrition, sensory organs, and other health topics, reflecting the diversity of topics encountered in clinical studies.

\section{Experimental Setup}

We conduct our experiments using a variety of large language models. The models were instructed to answer the question (thus producing a long answer) and also predict a final label (one of the S/R/NEI). For evaluation, predicted labels were collected and evaluated using classification metrics accuracy and macro-averaged F1 score, while the long answers were compared to the ground truth authors' conclusions with NLG evaluation metric ROUGE \cite{lin-2004-rouge}. Since this metric focuses on lexical overlaps, we also use the BERTScore metric, which captures semantic similarity \cite{zhang2020BERTScore}.

We chose three general-purpose LLMs and three biomedical LLMs. As a commercial state-of-the-art \ac{llm}, we include \textbf{GPT-4} (Turbo-0125) \cite{Achiam2023GPT4TR} in our comparison since it has demonstrated remarkable zero-shot performance on various \ac{nlp} tasks. Consequently, it is often used as a benchmark for comparing \acp{llm}. We also include two recent open-source models that achieved impressive performance, namely the \textbf{Mistral-7B} (Instruct-v2) \cite{jiang2023mistral}; and \textbf{Mixtral} \cite{jiang2024mixtral}, based on a sparse mixture-of-experts architecture \cite{fedus2022review}.

We found three domain-specific biomedical models that fit our use case: (1) \textbf{PMC-LLaMa 13B} \cite{wu2023pmcllama}, which is an extension of LLaMa \cite{touvron2023llama} additionally pre-trained on biomedical research articles from the database PubMed \cite{canese2013pubmed} and later fine-tuned on various datasets; (2) \textbf{MedAlpaca 7B} \cite{han2023medalpaca}, a biomedical extension of Alpaca (instruction-tuned version of LLaMa, \citeauthor{alpaca},\,\citeyear{alpaca}) fine-tuned on medical texts, encompassing resources such as medical flashcards, wikis, and dialogue datasets; and (3) \textbf{ChatDoctor (7B)} \cite{li2023chatdoctor}, another extension of LLaMa fine-tuned on patient-physician conversation datasets. 

Since the models were instruction-tuned with different templates, we used for each of them the base prompt reported by authors in technical reports. The prompt additionally instructed the models to predict one of the three classes and write an explanation for its answer (which we deem the "long answer"). All prompts can be found in Table~\ref{tab:llm_prompts}. GPT-4 was prompted through the OpenAI API, and Mixtral through the FastChat API\footnote{\href{https://github.com/lm-sys/FastChat}{FastChat: https://github.com/lm-sys/FastChat}} platform, which replicates OpenAI's chat completion API endpoint. The other four models were small enough to be loaded onto a single A100 GPU w/ 80 GB VRAM, and took around two computation hours each to generate all 2786 answers. We set the token limit to 256 and the temperature parameter to 0 for GPT-4 and Mixtral, maximizing deterministic generation by favoring high-probability words.

All the models were tested only in a zero-shot setting with just the question provided. Providing the full text of the abstract would make the experiments focus on \textit{machine reading comprehension}, which is not the goal of our study. We are interested in medical knowledge recall, which is best evaluated with a zero-shot setup. Initially, we also experimented with few-shot learning, but it only slightly helped in learning the formatting of answers and did not affect F1 performance since answers are still dependent on internal knowledge.

\section{Results and Discussion}

The performance metrics and results are summarized in Table~\ref{tab:performance-results}. The three general-purpose models performed better on average, both in the classification and the generation task. To our surprise, GPT-4 performed worse than expected and was beaten in classification by the smaller Mistral and MedAlpaca models, as well as Mixtral. In total, Mixtral exhibited the best performance.

\begin{table}[htpb]
\centering

\begin{tabular}{l|cc|cc}
\hline
           & \textbf{Acc}  & \textbf{F1} & \textbf{R-L } & \textbf{BSc}  \\ \hline
\textbf{Mistral} & 50.8 & 32.5 & 20.3 & 84.5 \\ 
\textbf{Mixtral}    & \textbf{62.0} & \textbf{34.8} & \textbf{21.1} & 85.6 \\ 
\textbf{GPT-4}      &   41.3   &   28.3   &   20.5   &   85.1   \\ \hdashline
\textbf{ChatDoctor} &   22.1   &  16.0    &   16.4   &    \textbf{86.0}  \\ 
\textbf{MedAlpaca}  &  55.2 &  33.4  &   19.3   &  85.4    \\ 
\textbf{PMC-LLaMa}  & 32.4 & 27.8 & 15.4 & 84.3 \\ \hline
\end{tabular}
%

\caption{\label{tab:performance-results} Performance results of six models, measured by accuracy (Acc) and F1-macro score for classification, Rouge-L (R-L), and BERTScore (BSc) for generation.}
\end{table}

The difference in final classification results is apparent in Table~\ref{tab:label-predictions} -- our dataset contains mostly NEI labels and the best-performing models tended to predict this label frequently. On the other hand, GPT-4 was very prone to answering questions affirmatively and saying they are supported by the latest medical research. Conversely, Mixtral was more reluctant to give definitive affirmative or negative verdicts, but correctly predicted in most cases that there is not enough reliable evidence to answer the question.  This is demonstrated in an example in Table~\ref{tab:nei_prediction}. When predicting \textsc{NEI}, the models did not refuse to answer but rather gave some background to the question and hints of positive or negative results in the past while finishing by saying the conclusive evidence is missing.

\begin{table}[htpb]
\centering

\resizebox{0.85\linewidth}{!}{%
\begin{tabular}{l|ccc}
\hline
           & \textbf{\textsc{Sup.}}  & \textbf{\textsc{Ref.}} & \textbf{\textsc{NEI}} \\ \hline
\textbf{Mistral} & 1277  &  30 & 1404 \\ 
\textbf{Mixtral} & 753 & 3 & 2030   \\ 
\textbf{GPT-4}      &   1835   &   5  &   946     \\ \hdashline
\textbf{ChatDoctor} &  2046 &  200  &   103   \\ 
\textbf{MedAlpaca}   &   1010  &  43    &   1733     \\ 
\textbf{PMC-LLaMa}  & 1417 & 682 & 687  \\ \hline
\textbf{Gold}  & \textbf{543} & \textbf{186} & \textbf{2057}  \\ \hline

\end{tabular}
}

\caption{\label{tab:label-predictions} Sum of predicted final labels (supported, refuted, not enough information) by each model.}
\end{table}

\begin{table*}[h]
\centering

\begin{tabular}{p{19mm}p{132mm}}
\hline
\textbf{Question} &  \small Do interventions to improve appropriate polypharmacy result in clinically significant improvement? \\[0.5ex]

\textbf{Conclusion}  & \small It is unclear whether interventions to improve appropriate polypharmacy resulted in clinically significant improvement. \cite{Cole2023-up} [\textbf{label}: \textcolor{blue}{\textbf{NOT ENOUGH INFO}}]  \\[0.5ex]

\hline
\textbf{GPT-4} & 
\small Interventions aimed at improving appropriate polypharmacy, which refers to the use of multiple medications in a manner that maximizes benefit and minimizes harm, have been shown to result in clinically significant improvements in various studies. These improvements include enhanced medication adherence, reduced adverse drug events, and better overall patient outcomes. [\textbf{predicted}: \textcolor{teal}{\textbf{SUPPORTED}}] \tikzxmark  \\[0.5ex]

\textbf{Mixtral} & 
\small While there is evidence suggesting that interventions can improve appropriate polypharmacy (the use of multiple medications for a patient), the impact on clinically significant improvement can vary depending on the specific patient population, interventions, and outcomes measured. Some studies have shown positive results, while others have not found significant improvements. Therefore, more research is needed to draw definitive conclusions. [\textbf{predicted}: \textcolor{blue}{\textbf{NOT ENOUGH INFO}}] \tikzcmark

    \\ \hline

\end{tabular}
\caption{Example of a question and conclusion from our dataset with generated answers by Mixtral and GPT-4 in a zero-shot setting. On average, GPT-4 produced more agreeable answers supporting the claims from the questions, while Mixtral on average had more critical answers and reported on uncertainty.}
\label{tab:nei_prediction}
\end{table*}

Among the three biomedical \acp{llm}, ChatDoctor had by far the worst performance and resorted to answering almost every question affirmatively (we observed this even when accounting for different values of hyperparameters \textit{temperature} or \textit{top p}). Interestingly enough, it was also the model with the highest BERTScore value of its generated text, showing that its fine-tuning on patient-doctor conversations most likely led to more authentic phrasing and word choices when delivering medical information. On the opposite side of the spectrum, PMC-LLaMa was the model that had by far the most \textsc{Refuted} predictions (which were basically nonexistent for GPT-4 and Mixtral). 

The disparity between \textsc{Refuted} and \textsc{NEI} predictions is a consequence of another common pattern we observed -- the models struggle to differentiate between these two classes. Questions labeled with \textsc{NEI} will usually have a conclusion saying there is not enough high-quality evidence to definitively answer the hypothesis. On the other hand, the \textsc{Refuted} questions will conclude that reliable studies show no effect of the healthcare intervention on the outcome, i.e., there is no difference between the tested intervention and a placebo drug or treatment. Since both of these classes contain "negative" phrasing, they were commonly mislabeled by models even when they were able to recall correct clinical evidence.

\begin{table}[htpb]
\centering
\resizebox{\linewidth}{!}{%
\begin{tabular}{l|ccc|ccc}

          \textbf{\# mentions} & \textbf{Mis.}  & \textbf{Mix.} & \textbf{GPT-4} & \textbf{CD} & \textbf{MA} & \textbf{PL}  \\ \hline
Cochrane & 321 & 105 & 21 & 2 & 3 & 52 \\  
review* & 759 & 663 & 390 & 8 & 614 & 477 \\  
meta-analys* & 810 & 306 & 214 & 587 & 75 & 35 \\  
\end{tabular}%
}

\caption{\label{tab:cochrane-mentions} No. of model responses mentioning the given term, showcasing its recall of relevant meta-reviews.}
\end{table}

Another interesting finding is shown in Table~\ref{tab:cochrane-mentions}. The models were prompted in a zero-shot setting with no reference to reviews, and yet in hundreds of answers, the models referred to reviews and meta-analyses related to the posed question, sometimes even referring to Cochrane itself. This was most evident in Mistral-7B, followed by Mixtral and MedAlpaca. This clearly demonstrates the encoded internal medical knowledge and the ability of models to refer to systematic reviews as the highest type of clinical evidence to answer given questions. 

Example response of all six LLMs for the same question is shown in Table~\ref{tab:six_answers}. Both Mistral and Mixtral referred to (different) systematic reviews on the given topic, while PMC-LLaMa referred to a specific randomized control trial. GPT-4 and MedAlpaca mentioned that the answer is based on clinical research, and ChatDoctor gave a plain affirmative answer. Still, it is evident that models tend to cite studies that are sometimes rather old. This can also lead to incorrect predictions and quoting outdated knowledge. An example is in Table~\ref{tab:old_knowledge}, where the systematic review from 2016 did not have enough information but the updated version of the same review from 2023 in our dataset did have enough new studies to give a positive verdict. How to update the outdated knowledge contained within LLMs is an ongoing challenge of \textit{knowledge editing} \cite{yao-etal-2023-editing, cohen2024evaluating}.

In the future, we see the dataset MedREQAL as a challenging testbed for medical knowledge recall in \acp{llm}, but also envision its potential for testing tasks like multi-document summarization, evidence retrieval, or retrieval-augmented generation.

\section{Conclusion }
We constructed a new biomedical QA dataset for testing the medical knowledge recall of \acp{llm}. The dataset originates from systematic reviews, synthesized evidence-based studies on research questions, a well-suited proxy for knowledge recall probing. We tested the performance of three general \acp{llm} and three biomedical \acp{llm}, showing that the scale of the model or its domain-tuning is not always tied to better performance. Deeper insights show that models have a moderately high level of recall and awareness of systematic reviews as quality evidence but still struggle with decisively concluding when there is not enough evidence to answer the question, leaving space for future improvements. 


\section*{Limitations}
Our dataset was built semi-automatically and relied a lot on using automated generative methods for its construction, which could have led to certain incorrect labeling. While the questions are usually just a simple interrogative rewording of the original objective and the long answers are original authors' conclusions, labels were completely generated from the given question and conclusion text. Our manual analysis of a 4\% subset of data showed that the performance is above 92\% correct. Keeping in mind that even human annotation is not perfect and always has inconsistencies in labeling, we deemed this a satisfying performance.

Our comparative analysis has certain limitations. We focus solely on a zero-shot setting and direct medical knowledge recall probing, and we acknowledge that settings of machine reading comprehension and in-context learning are also worthy of exploring in the future. We do not benchmark all relevant biomedical LLMs; some, like Med-PaLM, were computationally too expensive for us to run. Our study also lacks human evaluation of generated model responses, which could have shed more light on the qualitative performance and user-friendliness of the answers.

\section*{Ethical Considerations}
Our dataset and experiments deal with the highly sensitive domain of healthcare and medical NLP. While we probe the models in a zero-shot setting to elicit their internal medical knowledge for our research purposes, this is not a recommended way of their usage by end users or patients. Some responses still contained hallucinations and misleading medical advice that should be taken with a grain of salt and always manually checked within reliable sources or consulted with medical professionals.

\section*{Acknowledgements}
We would like to thank the anonymous reviewers for their helpful suggestions. This research has been supported by the German Federal Ministry of Education and Research (BMBF) grant 01IS17049 Software Campus 2.0 (TU München).

\bibliography{anthology,custom}

\newpage
\onecolumn
\appendix
\section{Appendix}
\label{sec:appendix-a}
The Appendix provides supplementary material about this study, including the model prompts in full length (Tables~\ref{tab:data_prompts} and \ref{tab:llm_prompts}) and example questions and model answers (Tables~\ref{tab:old_knowledge} and \ref{tab:six_answers}).

\begin{table*}[h]
\centering
\begin{tabular}{p{28mm}p{122mm}}
\hline
\textbf{Use Case} & \textbf{Prompt Content}\\
\hline
Classification & \verb|SYSTEM:| Your task is to classify an input with a medial question into one of several medical classes. If none of the classes fits, output miscellaneous. Output only one class from the following options:
\newline
cancer
\newline
cardiovascular system
\newline
dental and oral health
\newline
digestive system and metabolism
\newline
endocrine system and hormones
\newline
immune system and infections
\newline
cognitive and mental health
\newline
musculoskeletal system and dermatology
\newline
nutrition and dietary supplements
\newline
respiratory system
\newline
sensory organs
\newline
sexual and reproductive health
\\
\hline

Question \& Label generation & \verb|SYSTEM:| You're a helpful assistant. Your task is to help with generating questions and labels in the medical and clinical domain.
\newline \verb|AGENT| You will be given an excerpt of an abstract of a clinical systematic review. Based on the given background, objectives, and author's conclusions, generate only ONE SINGLE question, answerable with yes/no/uncertain, that sums up the main medical objective that was investigated. Please keep the question short and general and use the "Objectives" section to construct the question. The question should be about a general medical hypothesis, not about this specific review.
\newline
    Afterwards, please also give a label for the author's conclusions. Label tries to answer the objective by looking at the conclusion. The label may be ONLY from one of the following three: (1) SUPPORTED; (2) REFUTED; (3) NOT ENOUGH INFORMATION. Do not try to make up a new label. Please only select the third label if not enough evidence was found to reach a verdict, not if certainty of the conclusion is low! Please aim to predict "SUPPORTED" or "REFUTED" even if certainty of these conclusions by authors is low or moderate.
\newline
    Please structure the output in two lines, as:
\newline    
    QUESTION: (question)
\newline
    LABEL: (label)
\newline
    The documents begins now.

    \\ \hline

\end{tabular}
\caption{Overview of applied prompts for data generation and annotation.}
\label{tab:data_prompts}
\end{table*}

\begin{table*}[htpb]
\centering
\begin{tabular}{p{25mm}p{125mm}}
\hline
          \textbf{Model} & \textbf{Prompt Content}  \\ \hline
\textbf{Mistral} & <s>[INST]  Based on your knowledge, please answer this clinical question only with SUPPORTED (if the question is supported by the clinical research) or REFUTED (if the hypothesis is refuted by the current clinical research) or NOT ENOUGH INFORMATION (if there is insufficient evidence for the question in current research). Please give your output in form of LABEL: (label) . Briefly explain your answer.
\newline
QUESTION: \{question\}
\newline
[/INST]

\\ \hline
\textbf{Mixtral, GPT-4} & \verb|SYSTEM| You are an AI assistant helping answer clinical and medical questions based on your best knowledge.
\newline
\verb|AGENT| Please answer this clinical question only with SUPPORTED (if the question is supported by the clinical research) or REFUTED (if the hypothesis is refuted by the current clinical research) or NOT ENOUGH INFORMATION (if there is insufficient evidence for the question in current research). Please give your output in form of LABEL: (label) . Briefly explain your answer.
\newline
QUESTION: \{question\}
\\  \hline
\textbf{ChatDoctor, MedAlpaca, PMC-LLaMa} &  Below is an instruction that describes a task, paired with an input that provides further context. Write a response that appropriately completes the request.
\newline
\#\#\# Instruction: Based on your best current knowledge, please answer the following medical question. If you think there is not enough evidence to answer, then say so. Please answer the question with "SUPPORTED" or "REFUTED" or "NOT ENOUGH INFORMATION". Briefly explain your answer.
\newline
\#\#\# Input: \{question\}

\#\#\# Response:

  \\ \hline

\end{tabular}
%

\caption{\label{tab:llm_prompts} Prompts used for each of the \acp{llm} in experiments.}
\end{table*}


\begin{table*}[h]
\centering
\begin{tabular}{p{30mm}p{121mm}}
\hline
\textbf{Question} &  Does sucrose effectively relieve pain from heel lance in neonates? \\[1ex]
\hline
\textbf{Mixtral} & 

 While some studies suggest that sucrose may help to reduce procedural pain in neonates, including heel lance procedures, the evidence is not conclusive. The \textbf{Cochrane review on this topic from 2016} states that "there is considerable uncertainty about the effects of sucrose on procedural pain in newborn infants." Therefore, more research is needed to definitively answer this question. \newline [predicted label: \textcolor{blue}{\textbf{NOT ENOUGH INFORMATION}}]  \newline

    \\ 
\textbf{Latest Review \newline (in the dataset)} \newline & We reviewed 55 studies that included a total of 6273 babies. Twenty-nine studies had full-term babies only, 22 had preterm babies only, and four had both full-term and preterm babies. 
(...) Main results -- Sucrose compared to the control treatment \textbf{probably reduces pain from single heel lances} in babies. (...) \cite{Yamada2023} [original label: \textcolor{teal}{\textbf{SUPPORTED}}]

    \\ \hline

\end{tabular}
\caption{Example of a question from our dataset with the original study conclusion and the generated answer by Mixtral. In this case, the model recalled a systematic review on the correct topic, but which is outdated and deprecated. This shows the challenging nature of updating and editing the knowledge encoded within LLMs.}
\label{tab:old_knowledge}
\end{table*}

\begin{table*}[h]
\centering
\small
\begin{tabular}{p{19mm}p{132mm}}
\hline
\textbf{Question} &  \small Do antidepressants show efficacy in treating depressive symptoms in adults with cancer?  \\[1ex]

\textbf{Conclusion}  & \small Despite the impact of depression on people with cancer, the available studies were few and of low quality. This review found a potential beneficial effect of antidepressants against placebo in depressed participants with cancer. \cite{Vita2023-ja}  [\textbf{label}: \textcolor{teal}{\textbf{Supported}}]  \\[0.5ex]

\hline
\textbf{GPT-4} & 
\small Antidepressants have been studied for their efficacy in treating depressive symptoms in adults with cancer. \textbf{Clinical research, including randomized controlled trials and systematic reviews,} has generally supported the use of antidepressants for this purpose. These studies have found that antidepressants can be effective in reducing depressive symptoms in cancer patients, improving their quality of life. However, the effectiveness can vary depending on the type of antidepressant, the type of cancer, and the individual patient's response to the medication. 
\\[0.7ex]

\textbf{Mixtral} & 
\small There is evidence to support the use of antidepressants in treating depressive symptoms in adults with cancer. \textbf{A 2015 meta-analysis} of 52 randomized controlled trials found that antidepressants were more effective than placebo in reducing depressive symptoms in adults with cancer \textbf{(Kissane et al., 2015)}. However, it is important to note that the efficacy may vary depending on the type of antidepressant and individual patient factors. Therefore, a careful evaluation and monitoring by healthcare professionals is necessary when prescribing antidepressants for adults with cancer.  \\[0.7ex]

\textbf{Mistral} & 
\small Numerous studies have demonstrated that antidepressants are effective in reducing depressive symptoms in adults with cancer. \textbf{A meta-analysis of 28 randomized controlled trials }involving over 3,000 patients found that antidepressants were significantly more effective than placebo in treating depressive symptoms in individuals with cancer \textbf{(Kanji et al., 2011)}. Additionally, \textbf{a systematic review and meta-analysis of 35 studies }found that antidepressants were effective in reducing depressive symptoms in individuals with cancer, and that selective serotonin reuptake inhibitors (SSRIs) were more effective than tricyclic antidepressants (TCAs) \textbf{(Cassidy et al., 2015)}.
\\[0.8ex] \hdashline

\textbf{MedAlpaca} & 
\small Based on the \textbf{available research}, antidepressants appear to be effective in treating depressive symptoms in adults with cancer. Multiple studies have found that antidepressants can reduce depressive symptoms and improve quality of life in this population.  \\[0.7ex]

 \textbf{PMC-LLaMa} & 
\small The use of antidepressants to treat depressive symptoms in adults with cancer is supported by \textbf{a single randomized controlled trial (RCT) and a number of observational studies}. \textbf{The RCT, which was conducted in the United Kingdom}, included 152 patients with advanced cancer who were receiving palliative care. Patients were randomized to receive up to 3 months of treatment with either amitriptyline, a tricyclic antidepressant (TCA), or a nonsteroidal anti-inflammatory drug (NSAID). The study found that patients who received amitriptyline reported fewer depressive symptoms and better quality of life at the end of the study period than those who received the NSAID. (...) \\[0.8ex]

\textbf{ChatDoctor} & 
\small Yes, antidepressants have been shown to be effective in treating depressive symptoms in adults with cancer.  \\[0.5ex]
 
    \hline

\end{tabular}
\caption{Example of a question and conclusion from our dataset with generated answers by all six benchmarked LLMs. \textbf{Bolded} text includes references to literature and studies that were recalled by the models. Mixtral and Mistral named concrete systematic reviews (high-level evidence), PMC-LLaMa refered to a single randomized controlled trial (moderate-level evidence), while GPT-4 and MedAlpaca mentioned the research work in general. All six LLMs correctly predicted the \textcolor{teal}{\textbf{Supported}} label. }
\label{tab:six_answers}
\end{table*}

\end{document}